\documentclass[preprint,11pt,numbered, sn-mathphys-num]{sn-jnl}
\usepackage{tabularx}
\usepackage[utf8]{inputenc}
\usepackage[T1]{fontenc}
\usepackage[sort&compress]{natbib}
\usepackage{fancyhdr}
\usepackage{amssymb,amstext,amsmath, amsthm}
\usepackage{pifont}
\usepackage{booktabs}
\usepackage{bm}
\usepackage{graphicx}
\usepackage{notoccite}
\usepackage{color}
\usepackage[labelformat=simple]{subcaption}
\usepackage{csquotes}
\usepackage{longtable}
\usepackage{array}
\newcolumntype{L}[1]{>{\raggedright\let\newline\\\arraybackslash\hspace{0pt}}p{#1}}
\usepackage{svg}
\usepackage[ruled,longend,linesnumbered]{algorithm2e}
\usepackage{hhline}
\usepackage{multirow}
\usepackage[frozencache,cachedir=minted-cache]{minted}
\let\oldnl\nl
\newcommand{\nonl}{\renewcommand{\nl}{\let\nl\oldnl}}
\newcommand{\Mod}[1]{\ \mathrm{mod}\ #1}
\DeclareMathOperator\Arg{Arg}
\newlength{\boxwidth}
\def\btheorem{\begin{theorem}}
\def\etheorem{\end{theorem}}
\def\blemma{\begin{lemma}}
\def\elemma{\end{lemma}}
\def\bproposition{\begin{proposition}}
\def\eproposition{\end{proposition}}
\def\bcorollary{\begin{corollary}}
\def\ecorollary{\end{corollary}}
\def\bdefinition{\begin{definition}}
\def\edefinition{\end{definition}}
\def\bexample{\begin{example}}
\def\eexample{\end{example}}
\def\bremark{\begin{remark}}
\def\eremark{\end{remark}}

\newcommand{\be}{\begin{equation*}}
\newcommand{\ee}{\end{equation*}}

\newcommand{\beq}{\begin{eqnarray*}}
\newcommand{\eeq}{\end{eqnarray*}}
\newcommand{\bem}{\begin{multline}}
\newcommand{\eem}{\end{multline}}
\newcommand{\ba}{\begin{align*}}
\newcommand{\ea}{\end{align*}}

\renewcommand{\figurename}{Figure}
\renewcommand{\tablename}{Table}

\newtheorem{theorem}{Theorem}
\newtheorem{definition}{Definition}
\newcolumntype{Y}{>{\centering\arraybackslash}X}

\begin{document}

\title[Article Title]{Advancing multivariate time series similarity assessment: an integrated computational approach}

\author*[1,2]{\fnm{Franck B. N.} \sur{Tonle}}
\email{ftonle@icipe.org}

\author[1,3]{\fnm{Henri E. Z.} \sur{Tonnang}}

\author[2]{\fnm{Milliam M. Z.} \sur{Ndadji}}

\author[2]{\fnm{Maurice T.} \sur{Tchendji}}

\author[2]{\fnm{Armand} \sur{Nzeukou}}

\author[1]{\fnm{Kennedy} \sur{Senagi}}

\author[4]{\fnm{Saliou} \sur{Niassy}}

\affil[1]{\orgname{International Centre of Insect Physiology and Ecology (icipe)}, \orgaddress{\city{Nairobi}, \country{Kenya}}}

\affil[2]{\orgdiv{Department of Mathematics and Computer Science}, \orgname{Faculty of Science, University of Dschang}, \orgaddress{\city{Dschang}, \country{Cameroon}}}

\affil[3]{\orgname{University of KwaZulu-Natal}, \orgdiv{School of Agricultural, Earth, and Environmental Sciences}, \orgaddress{\country{South Africa}}}

\affil[4]{\orgname{African Union Inter-African Phytosanitary Council (AU-IAPSC)}, \orgaddress{\city{Yaoundé}, \country{Cameroon}}}

\abstract{
Data mining, particularly the analysis of multivariate time series data, plays a crucial role in extracting insights from complex systems and supporting informed decision-making across diverse domains. However, assessing the similarity of multivariate time series data presents several challenges, including dealing with large datasets, addressing temporal misalignments, and the need for efficient and comprehensive analytical frameworks. To address all these challenges, we propose a novel integrated computational approach known as Multivariate Time series Alignment and Similarity Assessment (MTASA). MTASA is built upon a hybrid methodology designed to optimize time series alignment, complemented by a multiprocessing engine that enhances the utilization of computational resources. This integrated approach comprises four key components, each addressing essential aspects of time series similarity assessment, thereby offering a comprehensive framework for analysis. MTASA is implemented as an open-source Python library with a user-friendly interface, making it accessible to both researchers and practitioners. To evaluate the effectiveness of MTASA, we conducted an empirical study focused on assessing agroecosystem similarity using real-world environmental data. The results from this study highlight MTASA’s superiority, achieving approximately 1.5 times greater accuracy and twice the speed compared to existing state-of-the-art integrated frameworks for multivariate time series similarity assessment. It is hoped that MTASA will significantly enhance the efficiency and accessibility of multivariate time series analysis, benefitting researchers and practitioners across various domains. Its capabilities in handling large datasets, addressing temporal misalignments, and delivering accurate results make MTASA a valuable tool for deriving insights and aiding decision-making processes in complex systems.}


\maketitle
\section{Introduction}
\label{sec:SectionI}
Data mining has evolved as a prominent area of research in computer science over the past few decades, enabling the discovery of knowledge from complex systems \citep{Krempl2014}. Many of these complex systems can be described through multivariate time series, which consist of multiple variables over multiple points in time, with measurements taken at regular time intervals \citep{Yin2014GeneralizedFF}. Time series analysis holds significant relevance and impact in various domains, including finance \citep{BOLLEN20111}, environmental sciences \citep{Alexander2007}, healthcare \citep{SZCZESNA2023102080, GUGGILAM2023102101}, and other fields \citep{Tran2022}. The profound insights and comprehensive understanding derived from the analysis of these time-dependent datasets hold the potential to greatly enhance informed decision-making and predictive modeling within these domains. Consequently, time series analysis has received increasing attention from researchers in recent years, emerging as one of the top ten most challenging problems in data mining \citep{Yang2006}. However, despite its significance and impact, time series analysis remains intricate and multifaceted, characterized by a multitude of unresolved issues \citep{Palpanas2019}. A prominent and commonly encountered challenge, which also serves as the focus of this paper, pertains to the evaluation of similarity in multivariate time series.

Assessing the similarity of time series holds tremendous value across a wide range of applications \citep{Morse2007, Xia2012}. An illustrative case can be found in the realm of intricate systems such as agroecosystems, where continuous monitoring of various variables such as temperature, precipitation, and wind speed results in multivariate time series. The capability to measure the similarity between such systems can significantly contribute to informed agricultural planning and management decision-making. It can help in predicting crop yield based on climatic patterns, identifying anomalous weather events, and pinpointing ecologically similar zones for optimizing resource allocation \citep{RamirezVillegas2011ClimateAF}.

Nevertheless, despite their widespread applications, conducting time series similarity assessments presents several challenges. Firstly, due to their considerable size, multivariate time series require a robust computational approach to ensure speed, efficiency, and scalability \citep{Jesson2015}. Secondly, time series often exhibit variations, noise, or temporal misalignments, which pose significant obstacles in achieving accurate similarity assessment \citep{Morse2007}. Time series alignment emerges as a critical challenge in this regard, requiring an efficient solution to address these issues. Another major challenge lies in the absence of comprehensive methodologies that seamlessly integrate all essential steps of similarity assessment (including feature representation, similarity measure, and similarity search) into a unified framework \citep{zheng2017survey}. This gap implies that researchers and practitioners seeking to assess time series similarity must possess a substantial understanding of the field \citep{Grinsted2004, Yin2014GeneralizedFF, Wang2014}. This represents a major obstacle to accessibility, considering the many fields of application derived from time series similarity.

In this article, we introduce an integrated approach for assessing the similarity between multivariate time series, referred to as Multivariate Time series Alignment and Similarity Assessment (MTASA). MTASA employs a hybrid methodology for time series alignment, incorporating key concepts from digital signal processing, including cross-correlation, convolution, and the discrete Fourier transform shifting theorem. This approach is enhanced by a multiprocessing engine that optimizes computational resources by efficiently distributing tasks among CPU cores.

The contributions of this article can be summarized as follows:
\begin{itemize}
    \item[--] Design of an integrated approach that addresses the three key steps of time series similarity assessment while emphasizing computational efficiency.
    \item[--] Development of a novel hybrid approach that harnesses fundamental concepts from digital signal processing to achieve precise alignment of time series data. 
    \item[--] Implementation of the proposed approach through the creation of an open-source Python library. This library aims to enhance accessibility to time series similarity assessment by providing a flexible, user-friendly tool for researchers and practitioners regardless of their expertise in time series analysis.
    \item[--] Comprehensive experimentation to empirically validate the effectiveness of MTASA on a large-scale, real-world multivariate time series dataset. The experimental study focuses on the assessment of agroecological similarity, effectively illustrating the practicality of MTASA in real-life scenarios.
\end{itemize}

The remainder of this article is organized as follows: Section \ref{sec:sectionII} presents an overview of related work. Sections \ref{sec:sectionIII} and \ref{sec:sectionIV} introduce the fundamental concepts of digital signal processing and problem formulation. Section \ref{sec:sectionV} outlines the proposed framework for assessing multivariate time series similarity. Section \ref{sec:sectionVI} presents an empirical evaluation of the approach to demonstrate its effectiveness and efficiency. Finally, in Section \ref{sec:sectionVII}, we draw conclusions from the findings.

\section{Related Work}
\label{sec:sectionII}

The assessment of multivariate time series similarity usually involves three key steps \citep{zheng2017survey}: 

\begin{itemize}
    \item[--] Feature representation: This initial stage involves transforming raw time series into a structure that captures the fundamental characteristics of the data, facilitating more effective similarity assessment \citep{Fakhrazari2017}. Feature representation techniques aim to reduce the dimensionality of the dataset, filter out noise, and retain the crucial patterns and structures relevant to similarity assessment.
    \item[--] Similarity measure: This step revolves around introducing novel metrics or methods to quantify the degree of similarity between time series entities, thereby influencing the accuracy of the similarity assessment \citep{Fakhrazari2017}. The Euclidean and Dynamic Time Warping (DTW) distances are two similarity measures commonly employed in time series analysis.
    \item[--] Similarity search: The final step entails identifying time series that exhibit similarity to a given sequence within a time series dataset, based on a selected similarity measure \citep{zheng2017survey}. The execution of a similarity search usually involves the concurrent execution of several tasks, namely indexing, normalization, clustering, and classification.
\end{itemize}

Although there is limited literature that integrates the aforementioned steps for time series similarity assessment, some research has explored this area. One notable pioneering study is by \citet{Grinsted2004}, which introduced a 4-step methodology based on the continuous wavelet transform, cross wavelet transform, phase angle, and wavelet coherence. This methodology analyzes and evaluates the similarity between two time series in the time-frequency domain, utilizing the Discrete Wavelet Transform (DWT) for feature representation and phase angle difference as a similarity measure. Similarity search is addressed using the wavelet coherence between the two-time series as a normalization method. The methodology is implemented through a Matlab software package, enabling users to efficiently perform cross-wavelet transform and wavelet coherence. However, this package lacks multiprocessing support, which can limit its scalability and efficiency when handling large datasets or complex computations. Furthermore, it is constrained to using DWT as the sole feature representation method, which restricts its adaptability in scenarios where other feature representations, such as the Discrete Fourier Transform (DFT), might be more appropriate.

In a similar vein, \citet{Wang2014} propose $S_{WBORDA}$, a methodology that combines dimensions for both whole sequence and subsequence matching similarity searches in multivariate time series datasets. The approach involves a sequential application of the principal component analysis (PCA), \textit{k} nearest neighbor (\textit{k}NN), and the weighted BORDA voting method. The BORDA voting method is a decision-making technique that consolidates results obtained from different similarity measures by assigning points to each candidate option based on their rank in individual preference lists. $S_{WBORDA}$ operates as follows: First, the multivariate data sequences are transformed into uncorrelated variables using PCA. Then, univariate time series similarity searching is conducted on each dimension. The top \textit{k} candidates are ranked and selected based on weighted BORDA scores, resulting in the final \textit{k}NN sequences for the query sequence. As for now, there is no associated software package available to automate or implement this approach.
 
Another notable study is presented in \cite{Yin2014GeneralizedFF}, which introduces a generalized framework for time series similarity assessment. The strength of this methodology lies in its generality, as it is founded on set theory, metric space, operator, and matrix theories, making it capable of supporting both univariate and multivariate time series as well as linear and nonlinear transformed time series. The framework consists of a set of mathematical definitions, theorems, and lemmas, laying a solid mathematical foundation for future research endeavors aimed at proposing innovative methodologies to various aspects of similarity assessment. Unfortunately, at present, there is no associated software or package available for this approach. In \cite{Lin2012RotationinvariantSI}, the authors proposed a similarity framework using a novel time series representation based on histograms, and recently, \cite{Liu2023} introduced another methodology centered around the use of spatial databases. The main challenge is that all these studies are theoretical, with no software or package available for their practical implementation. In contrast, there are a few studies that provide implementations of their proposed frameworks. These include CCAFS \citep{RamirezVillegas2011ClimateAF}, based on the Euclidean distance combined with a multidimensional Discrete Fourier Transform and implemented through an open-source R-package. The second approach, developed by \citet{GUIMAPI2022e02056}, employs rule-based modeling \citep{Liebhold2008} for similarity measure and search and is also implemented using the R programming language. 

It is worth noting that the existing works providing implementations of their proposed frameworks often heavily rely on domain-specific knowledge and assumptions, making them require significant adaptations to be effectively applied in different contexts. This observation underscores the importance of developing more generic methodologies that can be concurrently applied across diverse domains. A promising step in this direction can be seen in \cite{Donges2015} with \textit{pyunicorn}, a Python package based on complex network theory concepts, encompassing spatial network measures, node-weighted statistics, and network surrogates. As highlighted by the authors, this package holds applicability in various domains spanning from climate network similarity to functional brain network similarity, representing progress towards more versatile and adaptable methodologies. 

Nevertheless, the current landscape of proposed solutions, including pyunicorn, still exhibits several limitations. These limitations encompass the need for domain-specific knowledge for efficient utilization (e.g., pyunicorn requires a background in complex network theory), a lack of multiprocessing capabilities to accelerate the computations, and insufficient consideration of potential misalignments and temporal shifts within the time series. In response to these challenges, the MTASA methodology and its associated Python package have been developed. Table \ref{tab:frameworks} provides a comprehensive comparison of the different similarity assessment frameworks, showcasing their capabilities, implemented package/library, and limitations.

\begin{table}[h!]
\caption{Comparison of Time Series Similarity Assessment Frameworks}\label{tab:frameworks}%
\footnotesize
\begin{tabularx}{\textwidth}{@{} >{\hsize=0.85\hsize}X >{\hsize=1.35\hsize}X >{\hsize=0.8\hsize}X >{\hsize=0.7\hsize}X >{\hsize=0.7\hsize}X >{\hsize=1.6\hsize}X@{}}
\toprule
\textbf{Framework} & \textbf{Concepts} & \textbf{Time Series Alignment} & \textbf{Package/ Library} & \textbf{Parallel} & \textbf{Limitations}\\
\midrule
\citet{Grinsted2004} & DWT, Phase, Coherence & Yes & Matlab & No & No multiprocessing; Limited to DWT. \\ \midrule
CCAFS \citep{RamirezVillegas2011ClimateAF} & Euclidean Distance, Multidimensional DFT & Yes & Yes (R) & No & Domain-specific (specifically designed for climatic similarity); Lack of multiprocessing support. \\ \midrule
\citet{Lin2012RotationinvariantSI} & Histogram-based Representation & Yes & No & No & No associated package/library; The approach does not support multivariate time series. \\ \midrule
$S_{WBORDA}$ \citep{Wang2014} & PCA, k-NN, Weighted BORDA & Yes & No & No & No associated package/library; The approach have limited performances for large datasets. \\ \midrule
\citet{Yin2014GeneralizedFF} & Set Theory, Metric Space, Operators, Matrix Theories & No & No & No & No associated package/library. \\ \midrule
\textit{pyunicorn} \citep{Donges2015} & Complex Network Theory & No & Yes (Python) & Yes & Requires knowledge on complex network theory for efficient use;  Lack of time series alignment support. \\ \midrule
\citet{GUIMAPI2022e02056} & Rule-based Modeling & No & Yes (R) & No & Domain-specific (specifically designed for climatic similarity); Lack of multiprocessing support. \\ \midrule
\citet{Liu2023} & Spatial Databases & No & No & No & Domain-specific (specifically designed for geography); No associated package/library. \\ \midrule
MTASA (Proposed approach) & DFT, DFT Shifting, Cross-Correlation, Convolution & Yes & Yes (Python) & Yes & Limited to DFT and DWT for feature representation. \\ 
\bottomrule
\end{tabularx}
\end{table}

\section{Preliminaries}
\label{sec:sectionIII}

Within the context of this study, digital signal processing (DSP) can be defined as the application of mathematical operations to represent and manipulate signals digitally. In this section, we formally define key concepts of DSP that will be consistently employed throughout our study. Specifically, these are the following: \textit{discrete-time signal, Discrete Fourier Transform (DFT), inverse DFT, DFT shifting theorem, convolution, and cross-correlation.} \\

\begin{definition}
A \textbf{discrete-time signal}, or \textbf{discrete signal}, is a sequence of numbers denoted as $\{x_n\}_{n \in \mathbb{N}}$, where n denotes the time index, and $x[n]$ denotes the value of the $n$-th element in the sequence.
\end{definition}

\begin{definition}
\label{def:dft}
The \textbf{discrete Fourier transform} of a signal $\{x_n\}_{n \in \mathbb{N}}$ with $n = 0,1, \cdots, N-1$ is a transformation that maps the sequence $\{x_n\}$ from the time domain to the frequency domain, resulting in a new sequence of $N$ complex numbers $\{X_m\}$ with $m = 0,1, \cdots, N-1$ given by:
\begin{equation}
    X[m] = \sum_{n=0}^{N-1} x[n] \cdot e^{-j \cdot 2\pi \cdot \frac{m}{N} \cdot n}
\end{equation}
where $j$ is the imaginary unit $j= \sqrt{-1}$. The signal $\{x_n\}$ can be recovered from $\{X_m\}$ using the \textbf{inverse discrete Fourier transform} ($iDFT$) given by:
\begin{equation}
\label{eq:idft}
    x[n] = \frac{1}{N} \cdot \sum_{m=0}^{N-1} X[m] \cdot e^{\: j \cdot 2\pi \cdot \frac{m}{N} \cdot n} 
\end{equation}
\end{definition}

\begin{theorem}
Given a discrete signal $\{x_n\}_{n \in \mathbb{N}}$, let $\{X_m\}_{m \in \mathbb{N}} = \mathcal{F} \left(\{x_n\}\right)$ stand for the DFT of $\{x_n\}$ and $\{\text{\~{x}}_n\}_{n \in \mathbb{N}} = \mathcal{F}^{-1} \left(\{X_m\}\right)$ be the $iDFT$ of $\{X_m\}$. We then have $\{x_n\} \equiv \{\text{\~{x}}_n\}$, or, equivalently:
\begin{equation}
\label{eq:dft-identity}
    \mathcal{F}^{-1} \left[ \mathcal{F} \left(\{x_n\}\right) \right] = \{x_n\}
\end{equation}
\end{theorem} 

As highlighted in \cite{Agrawal1993}, the significance of the DFT within the context of time series analysis arises from the Fast Fourier Transform (FFT). This algorithm efficiently computes the sequence $\{X_m\}$, known as DFT coefficients, with a time complexity of $O\left(n \log n\right)$. For various signal operations (alignments, indexing), the FFT can reduce the time complexity from $O\left(n^2\right)$ to $O\left(n \log n\right)$. Another critical property mentioned in \cite{Agrawal1993} is that, for most practical sequences, using a few DFT coefficients (especially the first ones) in signal operations yields satisfactory performances comparable to using all the DFT coefficients. The efficiency of operations leveraging this property scales proportionally with the number and length of sequences, resulting, in certain contexts, in a reduction of time complexity from $O\left(n^3\right)$ to approximately $O\left(n \log n\right)$.\\

\begin{theorem}
\label{theorem:dft-shift}
A circular shift in the time-domain results in a phase shift in the frequency domain.

Consider a sequence $\{x_{n}^{d}\}_{n \in \mathbb{N}}$, obtained by a circular right shift of d elements from a periodic discrete signal $\{x_n\}_{n \in \mathbb{N}}$, as $x^{d} [n] = x[n + d]$. The DFT coefficients $\{X_{m}^{shifted}\}_{m \in \mathbb{N}}$ of the shifted signal $\{x_{n}^{d}\}$ are defined as follows: 
\begin{equation}
\label{eq:dft-shift}
    X^{shifted} [m] = e^{\: j \cdot 2\pi \cdot \frac{m}{N} \cdot d} \cdot X[m]
\end{equation}
\end{theorem}

This theorem signifies that a right shift of $d$ elements corresponds to a phase shift of $2\pi d m/N$ radians for the $m$-th DFT coefficient for a periodic discrete signal. The DFT shifting theorem holds practical utility as it allows us to determine the number of shifts required to align two signals by adjusting the phases of their DFT coefficients. Consequently, this theorem will be essential to our time series alignment logic.

\begin{definition}
Given two discrete signals $\{x_n\}_{n \in \mathbb{Z}}$ and $\{h_n\}_{n \in \mathbb{Z}}$ of length $N$, their \textbf{circular convolution}, denoted as $y = h \circledast x$ is a discrete signal of length $N$ defined as:
\begin{equation}
    y[k] = \sum_{n=0}^{N-1} h[n] \cdot x[- n + k \mod N ]
\end{equation}
The \textbf{cross-correlation} $z = h \star x$ between these is another discrete signal of length $N$ defined as:
\begin{equation}
\label{eq:cross-correlation}
    z[k] = \sum_{n=0}^{N-1} h[n] \cdot x[n + k \mod N ]
\end{equation}
\end{definition}

\begin{theorem}
Consider $\{x_n\}_{n \in \mathbb{Z}}$ and $\{h_n\}_{n \in \mathbb{Z}}$ two discrete signals of length $N$ and $y = h \circledast x$ the circular convolution between them. 

The DFT of the convolution is equal to the production of the DFTs:
\begin{equation}
    y = h \circledast x \Leftrightarrow Y[m] = H[m] \cdot X[m]
\end{equation}
\end{theorem}

Both the convolution and cross-correlation between two signals enable the examination of different shifting combinations to identify the optimal alignment in $O(n^2)$. The convolution theorem further facilitates the process by enabling faster computation in $O(n \log n)$.

\section{Problem formulation}
\label{sec:sectionIV}

To precisely formulate the problem, we need to formally define the following key concepts: \textit{multivariate time series dataset}, \textit{query sequence}, \textit{list of weights}, \textit{analysis period}, and \textit{similarity index matrix}.\\

\begin{definition}
\label{def:time-series-dataset}
A \textbf{multivariate time series dataset $\mathcal{T}$} is a collection of $K$ multivariate time series instances. Each multivariate time series consists of $N$ sequential measurements of $M$ distinct variables, defining the dataset as $\mathbf{T}^{\: K \times N \times M}$.
\end{definition}

\begin{definition}
\label{def:query-sequence}
A \textbf{query sequence $\mathcal{Q}$} is a multivariate time series represented as a matrix consisting of $N$ sequential measurements of $M$ different variables. Such sequence is defined as $\mathbf{Q}^{\: N \times M}$.\\
\end{definition}

By analogy with digital signal processing, a query sequence can be considered as a $1D$ array of length $M$, composed of discrete signals $\{x_n\}$ representing the sequential measurements. Similarly, a multivariate time series dataset can be perceived as a $K \times M$ matrix composed of discrete signals of length $N$.\\

\begin{definition}
\label{def:list-weights}
A \textbf{list of weights $\mathcal{W}$} is a $1D$ array consisting of $M$ values representing the importance of the different variables. This list is defined as $\mathbf{W}^{M}$.\\
\end{definition} 

The list of weights aims to facilitate scaling and normalization operations based on the significance of each variable. Various approaches, such as PCA, analytic hierarchy process, correlation analysis, or even expert knowledge, can be used to assign these weights. It is also assumed that the sum of the weights must be equal to $1$ to ensure relative proportionality across the variables, guaranteeing their adequate weighting and appropriate contributions to the overall system.\\

\begin{definition}
\label{def:analysis-period}
A \textbf{period of analysis $\mathcal{P}$} is a $1D$ array consisting of $n$ values representing the time indices of the measurement period of interest, defined as $\mathbf{P}^n$ with $1 \leq n \leq N$. \\
\end{definition} 

The analysis period allows the assessment of time series similarity for a specific period within the total duration of measurements. Defining an analysis period allows the evaluation to focus on a predefined time interval, which can be particularly useful in detecting potential temporal similarity patterns and identifying similarities that occur only during specific periods. \\

\begin{definition}
\label{def:sim-index-mat}
A \textbf{similarity index matrix $\mathcal{S}$} between $\mathcal{Q}$ and $\mathcal{T}$, denoted as $\mathbf{S}^{2 \times K}$, is defined as a $2 \times K$ matrix, wherein:
\begin{itemize}
    \item[--] $\mathbf{S}_{1,k} \in \{0, 1, \cdots, N-1\}, \forall k \in \{1, 2, \cdots, K\}$, represents the number of rotations required to correct misalignments or temporal shifts between $\mathcal{Q}$ and the $k$-th element of $\mathcal{T}$. The set $\{\mathbf{S}_{1,k}\}$ constituting the first row of $\mathcal{S}$ is termed as the $rotation\mathcal{A}rray$.
    \item[--] $\mathbf{S}_{2,k} \in [0, 1], \forall k \in \{1, 2, \cdots, K\}$ represents the normalized similarity (according to a specified similarity measure) between $\mathcal{Q}$ and the $k$-th element of $\mathcal{T}$ after a rotation of $\mathbf{S}_{1, k}$ elements has been applied. The set $\{\mathbf{S}_{2,k}\}$ constituting the second row of $\mathcal{S}$ is termed as the $similarity\mathcal{A}rray$. 
\end{itemize}
\end{definition} 

The similarity index matrix, as defined, provides a generic and intuitive measure of alignments and similarity between $\mathcal{Q}$ and the time series composing $\mathcal{T}$. By delivering comprehensive insights into the similarity relationship among $\mathcal{Q}$ and $\mathcal{T}$, this matrix serves as an effective output for the overall times series similarity assessment process described in Section \ref{sec:sectionII}. \\

Considering the previously defined concepts, the problem we seek to address can be precisely defined as:

\textbf{Given}: A multivariate time series dataset $\mathbf{T}^{K \times N \times M}$, a query sequence $\mathbf{Q}^{N \times M}$, a list of weights $\mathbf{W}^M$, and a period of analysis $\mathbf{P}^n$.

\textbf{Do}: Compute the similarity index matrix $\mathcal{S}$ between $\mathcal{Q}$ and $\mathcal{T}$ based on the specified list of weights $\mathcal{W}$ and the analysis period $\mathcal{P}$.\\

Efficiently solving this problem necessitates the consecutive application of fundamental steps inherent to time series similarity assessment, including feature representation, similarity measure, and search. Consequently, the problem that we aim to tackle in this paper can be reformulated as follows: \textit{Given a query sequence $\mathcal{Q}$, a multivariate time series dataset $\mathcal{T}$, a list of weights $\mathcal{W}$ and an analysis period $\mathcal{P}$, the goal is to design and implement an integrated multiprocessing time series similarity assessment approach, to efficiently compute the similarity index matrix between the query and the dataset, according to the list of weights and the analysis period.}

\begin{figure}
\centering
\includegraphics[width=\textwidth]{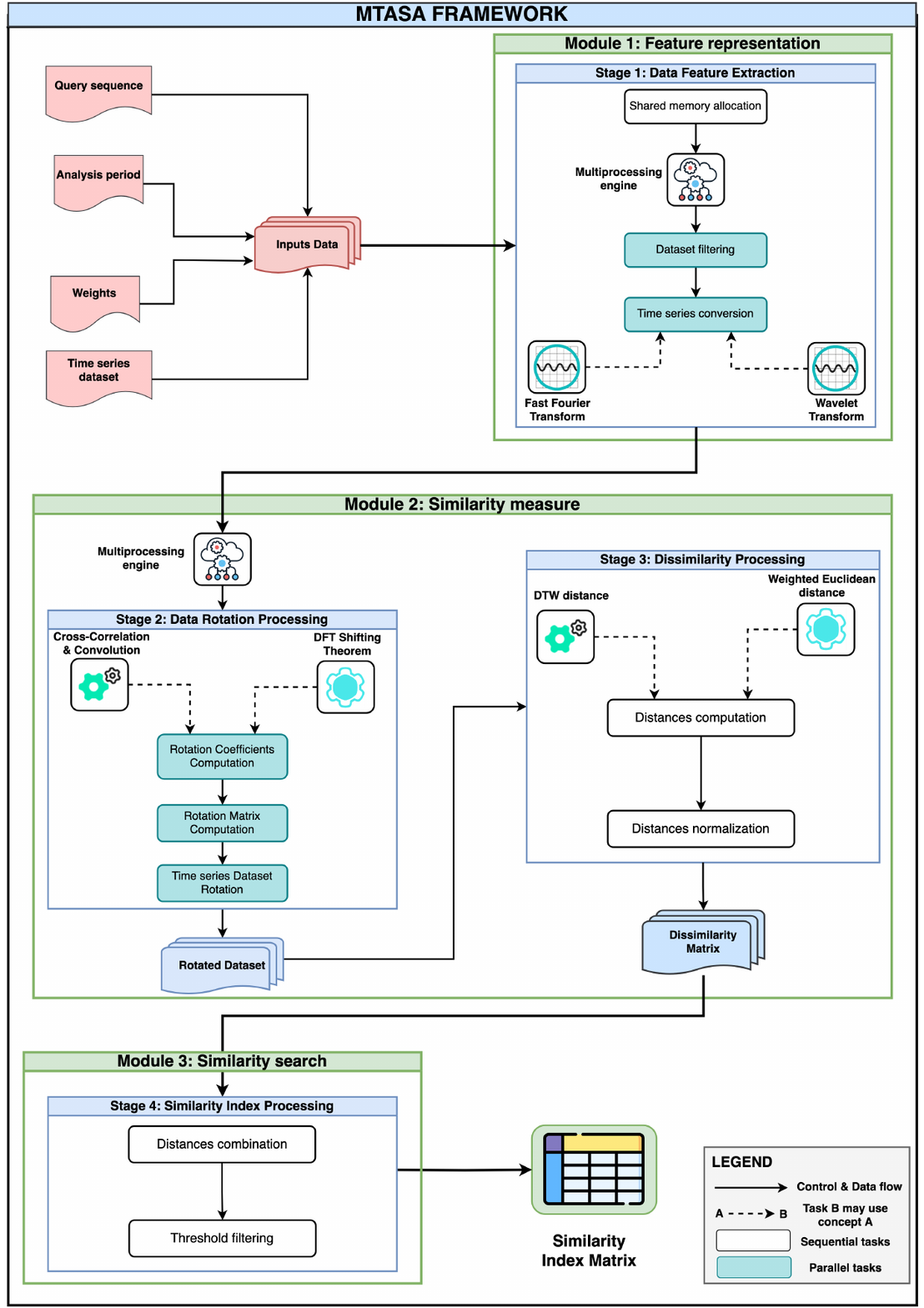}
\caption{The overall architecture of MTASA.}\label{fig:MTASA-framework}
\end{figure}

\section{Proposed approach}
\label{sec:sectionV}

In this section, we outline the proposed Multivariate Time series Alignment and Similarity Assessment (MTASA) approach designed to compute the similarity index matrix $\mathcal{S}$. MTASA consists of four primary stages:
\begin{itemize}
    \item[--] \textit{Data feature extraction}: The first stage involves preprocessing and feature extraction of both the query sequence and dataset. This is achieved by filtering and converting them into frequencies using either the DFT or DWT.
    \item[--] \textit{Data rotation processing}: In the second stage, the multivariate time series dataset $\mathcal{T}$ undergoes a rotation operation. This rotation is performed using a rotation matrix derived through the application of convolution, cross-correlation, and the DFT shifting theorem to the transformed frequencies.
    \item[--] \textit{Dissimilarity processing}: The third stage involves computing a dissimilarity matrix. This matrix is generated by evaluating and normalizing the similarity measure between the query sequence and each time series within the rotated dataset.
    \item[--] \textit{Similarity index processing}: In the final stage, the distances within the dissimilarity matrix are combined and filtered to produce the similarity index matrix, which provides a comprehensive assessment of similarity and alignment.
\end{itemize}

The overall framework of MTASA is presented in Fig. \ref{fig:MTASA-framework}. In the following paragraphs, we delve into each stage, explaining the methodologies and algorithms employed in MTASA.

\subsection{Data feature extraction}

In the data feature extraction phase, as outlined in Algorithm \ref{alg:mtasa} and Algorithm \ref{alg:rotation-mtasa}, the main objective is to perform feature representation within the MTASA framework. The process begins by allocating and assigning shared memory, referred to as $\mathcal{Q}_{shared}$. This shared memory will be accessed concurrently by multiple processors to handle the query sequence (see line 1, Alg. \ref{alg:mtasa}). 

The next step involves invoking a multi-processing engine (see line 3, Alg. \ref{alg:mtasa}) to facilitate the creation and management of a pool of processors for parallel execution of subsequent steps. The first of these steps is the filtration of the time series dataset into $\mathcal{T}_{valid}$ (see line 4, Alg. \ref{alg:mtasa}). This operation involves selecting valid multivariate time series, defined as those where measurements for all variables are consistently available throughout the entire measurement period. Time series with missing data are systematically excluded from the dataset, ensuring that only complete time series are retained for subsequent analysis. The following parallel step is the conversion of the valid time series into their frequency domain representations, following Definition \ref{def:dft}, while considering both the period of analysis and the rotation variables \footnote{The rotation variables can be defined as the measurement variables considered by the data rotation processor to correct temporal shifts between the query and the times series instances.} (see lines 5-7, Alg. \ref{alg:mtasa}; lines 1-12, 14-24, Alg. \ref{alg:rotation-mtasa}). This conversion can be performed using either the DFT or the DWT. The choice between these techniques depends on specific requirements: the DFT is suitable for a comprehensive, global frequency domain representation, whereas the DWT is preferable when a more localized, scale-based frequency analysis is desired \citep{Mrchen2003TimeSF}. The frequency domain representations are computed individually for each rotation variable and then combined by summation along the rows, resulting in a 1D array of length $\mid \mathcal{P}^n \mid$, which denotes the number of measurement steps in the analysis period. 

\begin{algorithm}[!htp]
\DontPrintSemicolon
\SetAlgoLined
\SetNoFillComment
\caption{MTASA - Compute Similarity Index Matrix}\label{alg:mtasa}
\KwData{$\mathcal{T} \: = \mathbf{T}^{K \times N \times M}$ - Multivariate time series dataset \tcp*[f]{see Def. \ref{def:time-series-dataset}} \\ \nonl \: \: \: 
          $\mathcal{Q} \ = \mathbf{Q}^{N \times M}$ \ \ \: - Query sequence \tcp*[f]{see Def. \ref{def:query-sequence}} \\ \nonl  
          \: \: \ $\mathcal{W} = \mathbf{W}^M$ \, \, \, \ \ \,  - List of weights \tcp*[f]{see Def. \ref{def:list-weights}} \\ \nonl  
          \: \: \:$\mathcal{P} \ = \mathbf{P}^n$ \: \: \: \: \ \ \ \ \ - Period of analysis \tcp*[f]{see Def. \ref{def:analysis-period}}
}
\KwResult{\: $\mathcal{S} \ \: = \mathbf{S}^{2 \times K}$ \: - Similarity index matrix between $\mathcal{Q}$ and $\mathcal{T}$, according to $\mathcal{W}$ and $\mathcal{P}$}
\tcc{Preprocessing, Feature extraction, and Rotated dataset computation}
$\mathcal{Q}_{shared} \gets \mathit{storeOnSharedMemory}(\mathcal{Q})$ \tcp*[f]{Share $\mathcal{Q}$ among all the CPU cores} \;
$rotation\mathcal{A}rray \gets createArray(K)$\;
\tcc{Multiprocessing computation}
$pool \gets createProcessPool()$\;
$(\mathcal{T}_{valid}, \mathcal{I}) \gets \mathit{getValidTimeSeries}(\mathcal{T})$\tcp*[f]{$k \in \mathcal{I} \Rightarrow \left( \forall n \leq N, \forall m \leq M, \ \mathbf{T}\left(k, n, m\right) \in \mathbb{R}\right)$} \;
\While{$\mathcal{T}_{valid} \neq \varnothing$}{
    $tasks \gets \left[\left(k, \mathbf{T}_{valid}\left(k\right), \mathcal{Q}_{shared}\right) \ for \ k \ in \ \mathcal{I}\right]$\;
    \tcc{Assign tasks to processes in pool}
    $results \gets mapToProcessPool(pool, \mathit{computeRotationCoef}, tasks)$ \tcp*[f]{See Alg. \ref{alg:rotation-mtasa}} \;
    \For{$(k, \mathit{rotationCoef}) \ in \ results$}{
        $rotation\mathcal{A}rray[k] \gets \mathit{rotationCoef} $\;
        $removeProcessedTimeSeries\left(\mathcal{T}_{valid}, \mathbf{T}_{valid}\left(k\right)\right) $\;
    }
 }
 $\mathcal{T}_{valid} \gets rotateDataset(\mathcal{T}, \ \mathcal{R}otation\mathcal{A}rray)$\;
$closeProcessPool(pool)$\;
\tcc{Dissimilarity matrix computation}
$dissimilarity\mathcal{M}atrix \gets createMatrix(K, M)$\;
\ForEach(\tcp*[f]{$\mathcal{M}$: list of measurement variables}){$m \ \in \ \mathcal{M}$}{
    $distances \gets computeSimilarityMeasure\left(\mathbf{T}_{valid}\left(\mathcal{P},m\right), \mathcal{Q}\left(\mathcal{P}, m\right)\right)$\;
    $results \gets normalizeDistances(distances)$\;
    $dissimilarity\mathcal{M}atrix[\_, m] \gets results * \mathbf{W}[m]$\;
}
\tcc{Similarity index matrix computation}
$similarity\mathcal{A}rray \gets combineDistances(dissimilarity\mathcal{M}atrix)$\;
$similarity\mathcal{A}rray \gets applyAbsoluteOrRelativeFiltering(similarity\mathcal{A}rray)$\; 
$\mathcal{S} \gets createMatrix(2, K)$\;
$\mathcal{S}(1, \_) \gets rotation\mathcal{A}rray$\;
$\mathcal{S}(2, \_) \gets similarity\mathcal{A}rray$\;
\Return $\mathcal{S}$
\end{algorithm}

\begin{algorithm}[!htp]
\DontPrintSemicolon
\SetAlgoLined
\SetNoFillComment
\caption{computeRotationCoef - Compute Rotation Coefficient}\label{alg:rotation-mtasa}
\KwData{$\mathbf{T}_k$ \: \: \: \: \: \: \: \: \: \: \ \ - Multivariate time series instance \\ \: \: \: \: \: \: 
          $\mathcal{Q} = \mathbf{Q}^{N \times M}$ \: \ \ \ - Query sequence\\ \: \: \: \: \: \:
          $\mathcal{P} = \mathbf{P}^n$ \: \: \: \: \: \ \ \ - Period of analysis\\ \: \: \: \: \: \:  
          $\mathcal{V} = \mathbf{V}^m$ \ \ \: \: \: \: \ \ - Indices of the rotation variables
}
\KwResult{$\mathit{rotationCoef} $ - Number of rotations required to correct temporal shifts between $\mathbf{T}_k$ and $\mathcal{Q}$}

\eIf{$ \mid \mathcal{P} \mid \ < N$}{
    \tcc{Subsequence similarity search}
    \eIf{$\mid \mathcal{V} \mid \ == 1$}{
        $\mathcal{Q}_{analysis} \gets \mathit{createArray}( \mid \mathcal{P} \mid )$\;
        $\mathcal{Q}_{analysis} \gets \mathcal{Q}\left( \mathcal{P}, \ \mathcal{V}  \right)$\;
        $X \gets DFTorDWT\left( \mathcal{Q}_{analysis} \right)$ \;
        $Y \gets DFTorDWT\left( \mathcal{T}_{k} \left(\mathcal{N}, \mathcal{V}\right) \right)$ \tcp*[f]{$\mathcal{N} = [0, 1, \cdots, N-1]$}
    } {
        $\mathcal{Q}_{analysis} \gets \mathit{createMatrix}\left(\mid \mathcal{P} \mid, \mid \mathcal{V} \mid\right)$\;
        $\mathcal{Q}_{analysis} \gets \mathcal{Q}\left(\mathcal{P}, \mathcal{V} \right) $\;
        $X \gets \sum DFTorDWT\left( \mathcal{Q}_{analysis} \right)$\;
        $Y \gets \sum DFTorDWT\left(  \mathbf{T}_{k} \left(\mathcal{N}, \mathcal{V}\right) \right)$\;
    }
    $\mathit{rotationCoef} \gets \left(N - \mathcal{A}rg\mathcal{M}ax\left( \Re \left( iDFT \left( \overline{Y} \circledast X \right) \right) \right)  \right) \Mod N $\;
} {
    \tcc{Full sequence similarity search}
    \eIf{$\mid \mathcal{V} \mid \ == 1$}{
        $\mathcal{Q}_{analysis} \gets \mathit{createArray}(N)$\;
        $\mathcal{Q}_{analysis} \gets \mathcal{Q}\left( \mathcal{N}, \ \mathcal{V}  \right)$\;
        $X \gets DFTorDWT\left( \mathcal{Q}_{analysis} \right)$\;
        $Y \gets DFTorDWT\left(  \mathbf{T}_{k} \left(\mathcal{N}, \mathcal{V}\right) \right)$\;
    } {
        $\mathcal{Q}_{analysis} \gets \mathit{createMatrix}\left(N, \ \mid \mathcal{V} \mid\right)$\;
        $\mathcal{Q}_{analysis} \gets \mathcal{Q}\left(\mathcal{N}, \mathcal{V} \right) $\;
        $X \gets \sum DFTorDWT\left( \mathcal{Q}_{analysis} \right)$\;
        $Y \gets \sum DFTorDWT\left(  \mathbf{T}_{k} \left(\mathcal{N}, \mathcal{V}\right) \right)$\;
    }
    $indexX_{max} \gets \mathcal{A}rg\mathcal{M}ax\left( \mid X[i] \ for \ i \in [1, N-1] \mid \right) $\;
    $indexY_{max} \gets \mathcal{A}rg\mathcal{M}ax\left( \mid Y[i] \ for \ i \in [1, N-1] \mid \right)  $\;
    \eIf{$indexX_{max} == indexY_{max}$}{
        $\theta \gets \Arg{ \left(X [indexX_{max}] \right)} - \Arg{ \left( Y [indexY_{max}] \right)} $\;
        $\mathit{rotationCoef} \gets \left( \theta \cdot \frac{N}{2 \cdot \pi \cdot indexX_{max}  } \right) \Mod N $\;
    } {
        $\mathit{rotationCoef} \gets \left(N - \mathcal{A}rg\mathcal{M}ax\left( \Re \left( iDFT \left( \overline{Y} \circledast X \right) \right) \right)  \right) \Mod N $\;
    }
}
\Return $\mathit{rotationCoef}$
\end{algorithm}

For each valid time series instance, the outcomes of the data feature extraction denoted as $X$ and $Y$, represent the summation of the frequency domain representations of the query sequence and the time series instance, respectively (see lines 5-6, 10-11, 18-19, 23-24, Alg. \ref{alg:rotation-mtasa}).

\subsection{Data rotation processing}
\label{sec:data-rotation}
Following the data feature extraction, the next stage in MTASA is the data rotation processing, which aligns the time series dataset with the query sequence, to facilitate the comparison of instances with temporal shifts (see lines 7-14, Alg. \ref{alg:mtasa}; lines 13, 26-35, Alg. \ref{alg:rotation-mtasa}). In this context, aligning these two elements is conceptualized as the process of synchronizing two discrete signals. 

To achieve this alignment, the first step is to compute the rotation coefficient between each valid dataset instance and the query sequence, then compile them into a $rotation\mathcal{A}rray$. Calculating the rotation coefficient involves determining the optimal number of shifts that minimize the disparity between two discrete signals. This is achieved by maximizing the dot product of the two series using cross-correlation \footnote{Cross-correlation is a widely used technique for DNA sequence alignment \citep{rockwood2005sequence, Brodzik2006}.} (see Equation \ref{eq:cross-correlation}). Consequently, the rotation coefficient between a time series instance $\mathbf{T}_{k}$ and the query sequence $\mathcal{Q}$ denotes the number of right shifts required to align each discrete signal within $\mathbf{T}_{k}$ with $\mathcal{Q}$. The computation of the rotation coefficient between $\mathcal{Q}$ and $\mathbf{T}_{k}$ for a given rotation variable $m$ can be expressed as follows:

Let's define a function called \textit{computeRotationCoefficient} that takes as inputs $\mathcal{Q}$, $\mathbf{T}_{k}$, and $m$. The output of this function, $rotation\mathcal{C}oef$, should satisfy the following property: 
\begin{equation}
	\forall c \in \{0, 1, 2, \cdots, N-1\}, \ z[c] \leq z[rotation\mathcal{C}oef]
\end{equation}
with $z$ a function of $c$, defined as: 
\begin{equation}
	z[c] = \sum_{n=0}^{N-1} \mathbf{Q}[n,m] \cdot \mathbf{T}[k, (n + c) \Mod N, m]
\end{equation}

which represents the cross-correlation between the elements of the query sequence and the time series instance for a particular rotation coefficient $c$.

Leveraging concepts from digital signal processing, we can efficiently ascertain the rotation coefficient between $\mathcal{Q}$ and $\mathbf{T}_{k}$ in logarithmic time complexity. Efficient computation of this coefficient relies on the frequency domain representation of $\mathcal{Q}$ and $\mathbf{T}_{k}$ denoted as $X$ and $Y$, respectively, which are obtained from the data feature extraction step. 

The determination of the rotation coefficient proceeds involves the following steps:

\begin{itemize}
    \item[--] Transform the cross-correlation into a convolution by inverting the time series instance in the time domain. This inversion is effectively achieved by conjugating the Discrete Fourier Transform (DFT) coefficients ${Y_n}$ within the frequency domain. 
    \item[--] Extract the real parts of the complex numbers obtained by applying the inverse DFT (using Equation \ref{eq:idft} and \ref{eq:dft-identity}) to the converted convolution.
    \item[--] Apply the ArgMax function, formally defined as $\mathcal{A}rg\mathcal{M}ax\left(A\right) = i \Rightarrow A[i] = max\left(A[j]\right), \forall j \in \{0, \cdots, |A|-1\}$. This function is applied to the real parts computed in the previous step to identify the number of rotations that maximize the convolution.
    \item[--] Convert this number into the rotation coefficient using a $modulo \ N$ operation on its additive inverse.
\end{itemize}

These four steps are summarized in Algorithm \ref{alg:rotation-mtasa} (see lines 13 and 32) using the following equation:
\begin{equation}
\begin{small}
\begin{aligned}
\mathit{rotationCoef} \gets \left(N - \mathcal{A}rg\mathcal{M}ax\left( \Re \left( iDFT \left( \overline{Y} \circledast X \right) \right) \right)  \right) \Mod N
\end{aligned}
\end{small}
\end{equation}

For the alignment of time series of length $N$, MTASA introduces a novel approach that combines these steps with the DFT shifting theorem. This integration capitalizes on the insight that the first DFT coefficients of highest amplitudes can be used for operations on discrete signals \citep{Agrawal1993}. In the context of full sequence similarity assessment, MTASA initiates the computation of the rotation coefficient by examining if the highest DFT coefficients of the two frequency domain representations ($X_{max}$ and $Y_{max}$) have the same analysis frequency. If they do, instead of composing multiple operations such as conjugation and convolution, the DFT shifting theorem (see Theorem \ref{theorem:dft-shift}) is used to determine the number of circular shifts needed to align the highest frequency components within $X$ and $Y$. This number of circular shifts serves as the rotation coefficient between $\mathcal{Q}$ and $\mathbf{T}_{k}$. Specifically, employing the DFT shifting theorem in this context involves calculating the difference of phase $\theta$ between the highest frequencies, followed by applying a $modulo \ N$ operation to the product of $\theta$ and the DFT shifting factor $2\pi d m/N$ (see Equation \ref{eq:dft-shift}). This process is summarized in Algorithm \ref{alg:rotation-mtasa} (see lines 26-30) through the following equation:

\begin{equation}
\mathit{rotationCoef} \gets \left( \theta \cdot \frac{N}{2 \cdot \pi \cdot indexX_{max}  } \right) \Mod N
\end{equation}

By integrating these two methodologies and leveraging its multiprocessing capabilities, MTASA efficiently computes the rotation coefficients for all valid time series instances simultaneously. These rotation coefficients are aggregated into an array called $rotation\mathcal{A}rray$ after computation. Each element of $rotation\mathcal{A}rray$ corresponds to a rotation applied to the respective time series instance, effectively aligning each instance with the query sequence and resulting in a rotated dataset serving as the output of the data rotation processing. 

\subsection{Dissimilarity processing}

Following the computation of the rotated time series dataset, MTASA proceeds to calculate the dissimilarity matrix. The dissimilarity processor performs this through a two-step process involving similarity measurement and distance normalization. The \textbf{dissimilarity matrix $\mathcal{D}$}, denoted as $\mathbf{D}^{K \times M}$, is defined as a $K \times M$ matrix, wherein $\mathbf{D}_{k,m}$ represents the weighted normalized distance between $\mathcal{Q}$ and the $k$-th element of $\mathcal{T}$ considering only the measurement variable $m$. \\

\textbf{1 - Computing the dissimilarity distance.} The computation of the dissimilarity matrix starts with the evaluation of the distance between each valid time series instance and the query sequence, a process completed for every measurement variable $m \in \mathcal{M}$. By default, MTASA employs the Euclidean distance (see Equation \ref{eq:euclidean}) as a similarity measure for this operation. This choice is justified because, by default, the data rotation processor has already applied an alignment process. Consequently, the Euclidean distance, which measures the "straight line" distance between two points in a multidimensional space, is relevant as it establishes a direct correspondence between the components constituting the two multidimensional points of interest, effectively exploiting alignment.

\begin{equation}
\label{eq:euclidean}
    EuclideanDistance(\mathbf{Q}, \mathbf{T}_{k}, m) = \sqrt{\sum_{i=1}^{N} \left(\mathcal{Q}[i,m] - \mathbf{T}[k,i,m] \right)^2}
\end{equation}

However, MTASA supports the use of the DTW distance (see Equation \ref{eq:dtw}) as an alternative to manage temporal misalignment if the data rotation processing has not been performed (line 17 in Algorithm \ref{alg:mtasa}). This addition enhances MTASA's flexibility to provide accurate similarity measures even in the absence of preliminary data alignment, catering to a broader range of use cases.

\begin{equation}
\label{eq:dtw}
\begin{small}
\begin{aligned}
DTW(\mathcal{Q}, \mathbf{T}_{k}, m) &= \mid \mathbf{Q}[0,m] - \mathbf{T}[k,0,m] \mid \\
&\quad + \min \begin{cases}
      DTW (\mathbf{Q}, \mathbf{T}_{k}[1:N-1 , m]) \\
      DTW (\mathbf{Q}[1:N-1 , m], \mathbf{T}_{k}) \\
      DTW (\mathbf{Q}[1:N-1 , m], \mathbf{T}_{k}[1:N-1 , m])
    \end{cases}
\end{aligned}
\end{small}
\end{equation}

\textbf{2 - Normalizing and weighting of computed distances.} After computing the dissimilarity distances, MTASA normalizes these values to simplify the subsequent weighting process. This process is achieved through a min-max normalization, which scales the distances within a range of $[0, 1]$. Normalizing the distances to this range aligns them with the weights of the measurement variables, which are also restricted to the interval $[0, 1]$ and sum up to $1$. Consequently, a direct and intuitive correspondence emerges between normalized distances and the weights of associated measurement variables. The weighting of the distances involves multiplying the normalized distances by the respective weights of the variables. The overall procedure for computing the dissimilarity matrix can be summarized using the following equation:

\begin{equation}
\begin{small}
\begin{aligned}
\mathcal{D}[k,m] = \frac{distance(Q, \mathbf{T}_{k}, m) - min\{distance\left(Q, \mathbf{T}, m\right)\}}{max\{distance\left(Q, \mathbf{T}, m\right)\} - min\{distance\left(Q, \mathbf{T}, m\right)\} } \cdot \mathbf{W}[m]
\end{aligned}
\end{small}
\end{equation}

\subsection{Similarity index processing}

The final step of MTASA is to compute and return the similarity index matrix $\mathcal{S}$. As previously defined, $\mathcal{S}$ consists of a $rotation\mathcal{A}rray$ and a $similarity\mathcal{A}rray$. The $rotation\mathcal{A}rray$ has already been computed as an intermediate result for the rotation of the time series dataset during the data rotation processing.

Given the computed dissimilarity matrix $\mathcal{D}$ computed in the previous step, the computation of the $similarity\mathcal{A}rray$ involves combining and optionally filtering the elements of $\mathcal{D}$. Since the distances have been normalized and weighed, the maximum value $\mathbf{D}_{k,m}$ for every column $m$ within $\mathcal{D}$ is less than or equal to $\mathbf{W}[m]$. The computation of $similarity\mathcal{A}rray$ consists of summing the elements of $\mathcal{D}$ along the columns, resulting in a $1D$ array containing values representing the similarity between the time series instances of $\mathcal{T}$ and $\mathcal{Q}$. These values are referred to as \textit{"similarity indices"}. Due to the weighting and the propriety that ($\sum_{m} \mathbf{W}_{m} = 1$), all similarity indices belong to the range $[0, 1]$. 
 
Additionally, MTASA supports optional filtering of the \\ $similarity\mathcal{A}rray$ based on specific similarity threshold criteria. Two types of filtering are supported: "absolute filtering", which removes similarity indices below a particular value, and "relative filtering", which selects the $k$ nearest neighbors of $\mathcal{Q}$ within $\mathcal{T}$ based on their similarity index. These filtering options accommodate scenarios where strict similarity thresholds or relative rankings of time series instances are essential.

To construct the similarity index matrix, the $rotation\mathcal{A}rray$ is assigned to the first row, and the $similarity\mathcal{A}rray$ to the second. $\mathcal{S}$ is then returned as the final output of the integrated MTASA approach.

\subsection{\textit{pymtasa}: A python implementation of MTASA}

The paper introduces an open-source Python package called \textit{"pymtasa"} which implements the MTASA approach. This package aims to make MTASA accessible to researchers and practitioners by providing a user-friendly interface that facilitates experimentation. \textit{pymtasa} follows the principles of object-oriented programming, resulting in a clear and coherent structure that enhances the maintainability and ensures a clear representation of the intricate components involved within MTASA. It leverages popular scientific computing libraries, especially numpy \citep{harris2020array}, to optimize numerical operations, accelerating the computation of dissimilarity and similarity measures. Additionally, numpy's compatibility with various data formats and its capabilities to handle large datasets make it an ideal choice for MTASA. By leveraging Python's established scientific computing libraries, \textit{pymtasa} represents a valid implementation of MTASA and serves as a vehicle to advance the understanding and adoption of this innovative approach across the data mining and scientific communities. An example of \textit{pymtasa} use is shown below\footnote{The complete source code and documentation can be found in a GitHub repository at \url{https://github.com/icipe-official/pymtasa}}. \\

    \begin{minted}{python}
from pymtasa.mtasa_configuration import MTASAConfiguration
from pymtasa.similarity import Similarity
from pymtasa.static_variables import RESULTS_DIRECTORY,TIME_SERIES_DATASET, 
QUERY_SEQUENCE

if __name__ == '__main__':
    mtasa_configuration = MTASAConfiguration(
        measurement_vars=("prec", "tmean"),
        weights=(0.5, 0.5),
        ref_data=QUERY_SEQUENCE,
        target_dataset=TIME_SERIES_DATASET,
        analysis_period=[11, 12],
        rotation_variables=["prec", "tmean"],
        threshold=0,
        rotation_mode=True,
        threshold_mode=True,
        outfile=RESULTS_DIRECTORY,
        file_name="pyresults",
        write_file=True
    )

    similarity = Similarity(mtasa_configuration)
    similarity_index_matrix = similarity.compute_similarity_matrix()
\end{minted}

\section{Application}
\label{sec:sectionVI}

To demonstrate its efficacy and versatility, we applied MTASA to a domain-specific problem: \textit{assessing the suitability of a region for an invasive pest based on its agroecological similarities to areas with abundant pest presence.} Several factors justify this problem selection, including the complex nature of agroecosystems characterized by extensive multivariate time series data, the need for domain expertise, and the challenges posed by temporal shifts among agroecosystems \citep{RamirezVillegas2011ClimateAF, Ordano2013}. The lack of labeled datasets also hinders the application of time series similarity assessment approaches based on AI models \citep{Tuia_2022}. In this regard, employing MTASA to address this challenge is a compelling demonstration of its potential in aiding researchers to assess time series similarity without an extensive expert background, simultaneously managing temporal shifts and ensuring computational efficiency.

\subsection{Application protocol}

Considering a multivariate time series dataset denoted as $\mathcal{T}$ wherein each instance represents the agroecology of a specific site, a predefined list of known sites with high pest population density denoted as $\mathcal{L}$, an analysis period $\mathcal{P}$ and a list of weights $\mathcal{W}$, our protocol include the following steps:

\begin{enumerate}
    \item \textbf{Reference site (query sequence) selection:} This process begins by selecting the site $s$ with the highest level of pest population density within $\mathcal{L}$. The corresponding time series for this selected site is denoted $\mathcal{Q}_s$.
    \item \textbf{Similarity computation and filtering:} After selecting site $s$, we compute the similarity index matrix $\mathcal{S}$ between $\mathcal{Q}_s$ and $\mathcal{T}$, according to $\mathcal{P}$ and $\mathcal{W}$. Then after, a filtering process is implemented on $\mathcal{S}$ to identify the sites in the dataset that are most similar to the reference site $s$ based on a predefined threshold, resulting in a set of similar sites denoted $\mathcal{S}_{sim}$.
    \item \textbf{Validation}: For each site $l$ within the list $\mathcal{L}$, it is determined whether $l$ is included in the set $\mathcal{S}_{sim}$. If $l$ is present in $\mathcal{S}_{sim}$, it indicates that MTASA has successfully assessed the similarity of $l$ to $s$, validating that the approach effectively recognizes sites with high pest species suitability.
    \item \textbf{Accuracy determination:} The accuracy of MTASA is quantified as the ratio between the number of correctly assessed sites (i.e., sites from $\mathcal{L}$ correctly identified as similar to $s$) and the total number of sites within $\mathcal{L}$.
    \item \textbf{Execution time measurement:} The computational efficiency of MTASA is evaluated by measuring the execution time required to perform the time series similarity assessment between the reference site $s$ and the dataset $\mathcal{T}$.
\end{enumerate}

In summary, the evaluation protocol involves selecting a reference site with high pest density, computing similarity indices, identifying the most similar sites, verifying the assessment accuracy, and measuring the execution time. The accuracy of MTASA is determined by its ability to correctly evaluate the suitability of sites within a dataset through a time series similarity assessment, using a high pest density site as the query sequence. Furthermore, the execution time measurement provides insights into the computational efficiency of MTASA in performing time series similarity assessments.

\begin{table}[h!]
\caption{MTASA parameters used for the experiment.}\label{tab:parameters}
\begin{tabularx}{\textwidth}{@{} l>{\hsize=13\hsize}X >{\hsize=17\hsize}X @{}}\hline
\textbf{Parameter} & \textbf{Value} & \textbf{Justification} \\ \midrule
Measurement period & May 2018 - December 2018 & Availability of data from the FAO and CHELSA BIOCLIM+ dataset.\\ \midrule
Measurement variables & Temperature, precipitation, wind speed, solar radiation & The 4 variables with the highest impact on FAW suitability \citep{GUIMAPI2022e02056}.\\ \midrule
Rotation variables & Temperature, precipitation & Variables most susceptible to temporal shifts \citep{RamirezVillegas2011ClimateAF}. \\ \midrule
Analysis period & May 2018 - July 2018 and November 2018 - December 2018 & Peak activity period of FAW \citep{GUIMAPI2022e02056}. \\ \midrule
List of weights & [0,35; 0,25; 0.2; 0.2] & Temperature and precipitation have a major influence on FAW lifecycle \citep{PaudelTimilsena2022}. \\ \midrule
Filtering and threshold & Relative filtering, based on threshold above the 3rd quartile of similarity values. & Same methodology was used in \citep{GUIMAPI2022e02056} to define the thresholds. \\
\bottomrule
\end{tabularx}
\end{table}

\subsection{Experimental setup}

The experimentation aimed to assess the suitability of the African continent for the Fall armyworm, a highly invasive insect pest species native to North America that has rapidly spread across Africa over the last seventh years. FAW poses a significant threat to maize, sorghum, and millet production in sub-Saharan Africa, causing annual losses of billions of dollars \citep{Tonle2024}. To evaluate the suitability of the African continent for FAW, we employed the MTASA. We selected a well-known site with a high population density of FAW for our analysis. MTASA was used to measure the similarity between this site and the broader African continent. We then assessed the accuracy and the execution time of MTASA according to the specified application protocol. For benchmarking purposes, we compared the performances of MTASA with two state-of-the-art approaches dedicated to the evaluation of agroecological similarity: CCAFS \citep{RamirezVillegas2011ClimateAF} based on the Euclidean distance combined with a multidimensional Discrete Fourier Transform and  Guimapi et al.'s methodology \citep{GUIMAPI2022e02056} based on rule-based modeling. We selected these two methodologies for their reliance on expert knowledge, ensuring robustness and precision in assessing similarity.

The experiment was run on a machine powered by a 3.90GHz Intel Xeon processor with 16 cores and 128 GB of RAM. FAW occurrence and density are known to depend heavily on four climatic variables: temperature, precipitation, wind speed, and solar radiation \citep{PaudelTimilsena2022, GUIMAPI2022e02056}. Therefore, we used a dataset containing $108,154,809$ time series instances, each providing monthly measurements of these four variables measured over 8-month period of interest. The dataset was sourced from the CHELSA BIOCLIM+ dataset \citep{bioclim_plus2022}, which offers high spatial resolution bioclimatic variables on global scale. The analysis period included May to July and November to December, corresponding to the peak activity of FAW during the maize cropping season. To determine the accuracy, we compiled a list of 2756 high pest population density sites from field data monitored by the United Nations Food and Agriculture Organization (FAO). For each month within the measurement period, we applied our application protocol using the site with the highest pest density for that respective month, according to the configuration presented in Table \ref{tab:parameters}.

\begin{table*}[h]
\centering
\caption{Monthly Performance Metrics for MTASA, CCAFS, and Guimapi et al. Methods}
\label{tab:results}
\begin{tabularx}{\textwidth}{l *{7}{Y}} 
\toprule
& & \multicolumn{2}{c}{MTASA} & \multicolumn{2}{c}{CCAFS} & \multicolumn{2}{c}{Guimapi et al.} \\
\cmidrule(l){3-4} \cmidrule(l){5-6} \cmidrule(l){7-8}
Month & Validation sites &  Accuracy (\%) & Time execution (s) & Accuracy (\%) & Time execution (s) & Accuracy (\%) & Time execution (s)\\
\midrule
May & 52 & 87 & 1285 & 37 & 2668 & 68 & 2883\\
June & 171 & 58 & 1351 & 22 & 2929 & 42 & 3147\\
July & 193 & 80 & 1367 & 59 & 2574 & 41 & 2868\\
August & 339 & 55 & 1354 & 18 & 2697 & 42 & 3032\\
September & 429 & 58 & 1371 & 27 & 2661 & 43 & 2978\\
October & 472 & 72 & 1286 & 43 & 2220 & 55 & 2677\\
November & 550 & 93 & 1394 & 68 & 2613 & 50 & 2912\\
December & 550 & 77 & 1312 & 24 & 2600 & 38 & 2887\\
\midrule
\midrule
Average & 346 & 72 & 1340 & 37 & 2620 & 47 & 2923\\
\bottomrule
\end{tabularx}
\end{table*}

\subsection{Results}

In Table \ref{tab:results}, we can delve deeper into the results of our experiment and gain valuable insights into the performance of the three assessed approaches: MTASA, CCAFS, and Guimapi et al.'s methodology. These insights provide a clear understanding of how each method performed in terms of accuracy and execution time, as outlined in our methodology.

Table \ref{tab:results} presents a monthly breakdown of performance metrics for the approaches. These metrics are crucial in assessing the effectiveness of each method in determining the suitability of the African continent for the Fall armyworm (FAW). Specifically, the metrics include accuracy percentages and time execution in seconds, all measured according to our established methodology. What stands out in the table is the consistent superiority of MTASA across all the months of the assessment. MTASA consistently outperforms the alternative methods, demonstrating its robustness and reliability. Notably, during the crucial months of May, November, and December, MTASA achieves remarkable accuracy levels of 87\%, 93\%, and 77\%, respectively. This outstanding performance of MTASA can be attributed to its unique capabilities. MTASA excels in efficiently managing temporal shifts, enabling it to capture intricate and dynamic patterns that exist within the query sequence and the time series instances. This proficiency is especially critical in assessing FAW suitability, where factors such as climate and environmental conditions are subject to temporal variations.

\begin{figure}[h!]
\centering
\includegraphics[width=0.85\textwidth]{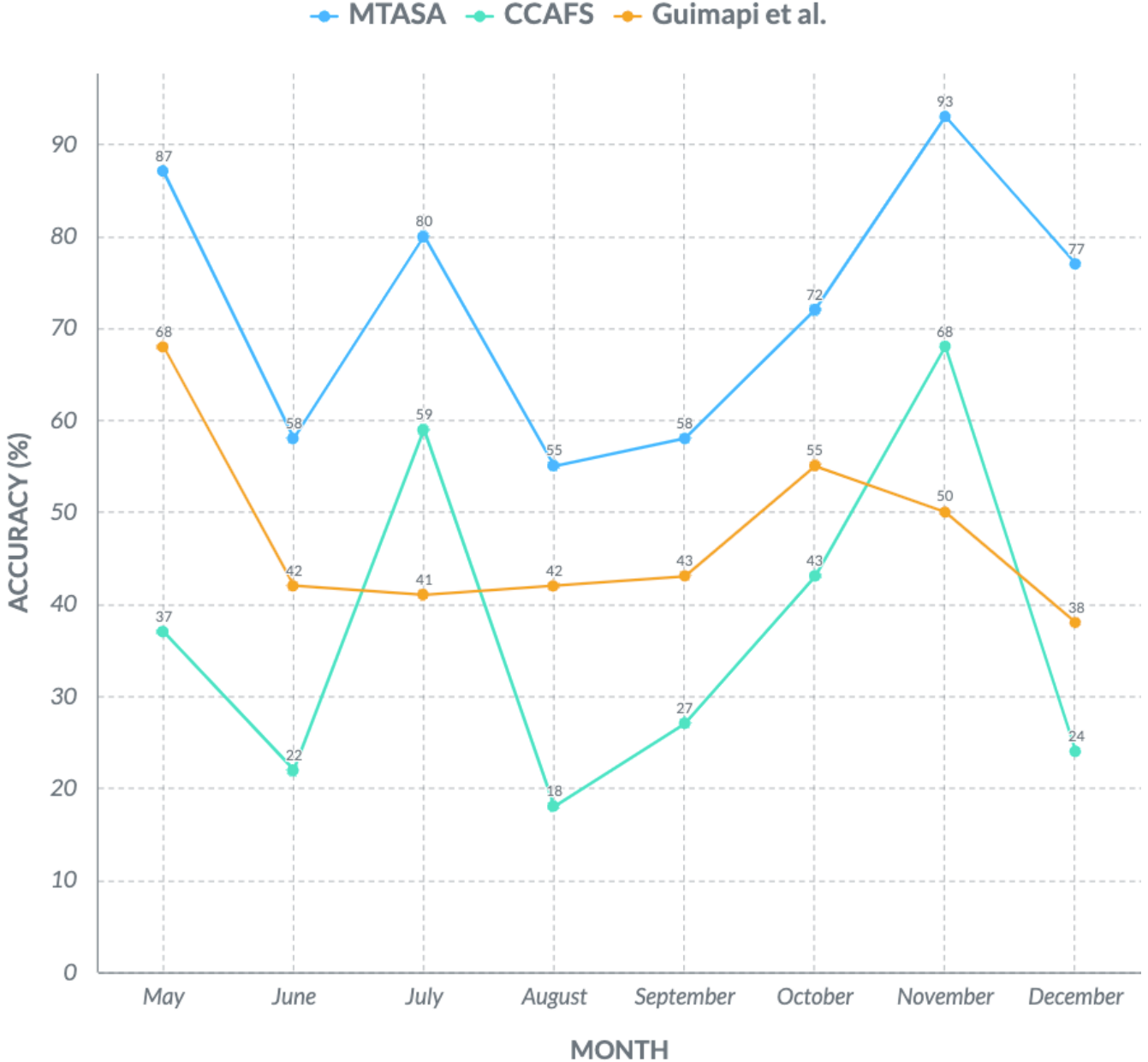}
\caption{Monthly accuracy percentages of MTASA and state-of-the-art approaches}\label{fig:mtasa-accuracy}
\end{figure}

Upon a thorough examination of the source code for CCAFS, we discovered a critical factor contributing to its lower accuracy performance. Specifically, the issue was related to the systematic use of the first Discrete Fourier Transform (DFT) coefficient for time series alignment. This limitation became apparent as we delved into the methodology and implementation of CCAFS. As detailed in section \ref{sec:data-rotation} of our study, this limitation in CCAFS led us to explore innovative solutions within MTASA to overcome this hurdle. MTASA introduced a groundbreaking hybrid approach that combines elements of cross-correlation, convolution, and DFT shifting. This synergistic integration of different techniques aimed to address the shortcomings observed in CCAFS and ultimately improve the alignment of data, resulting in enhanced overall performance. The use of this hybrid approach in MTASA brought about notable improvements in accuracy when compared to CCAFS. By leveraging cross-correlation, convolution, and DFT shifting together, MTASA achieved superior data alignment. This alignment is crucial in accurately capturing the intricate patterns and relationships among the data, especially in the context of assessing the suitability of the African continent for the Fall armyworm.

\begin{figure}[h!]
\centering
\includegraphics[width=0.85\textwidth]{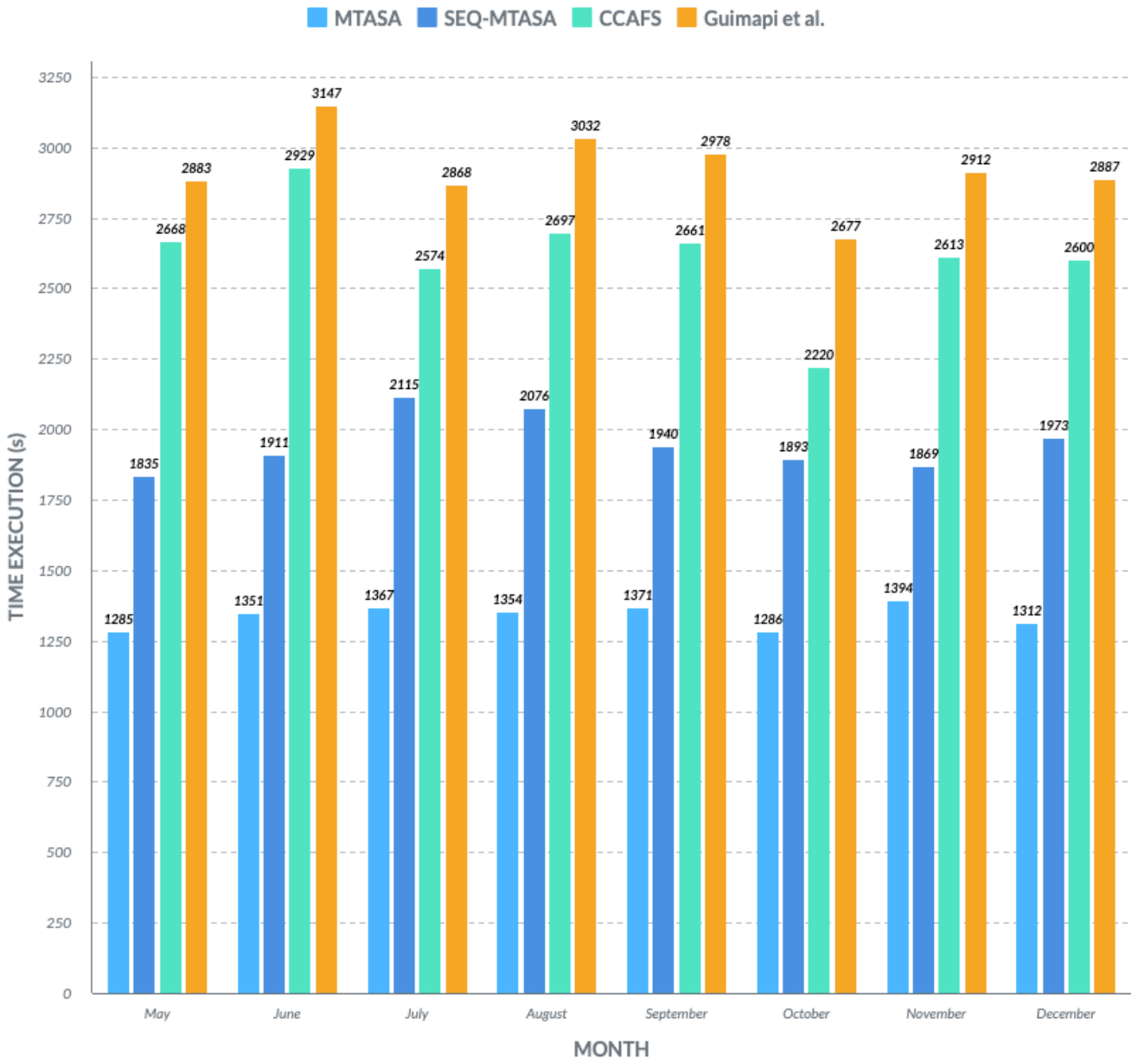}
\caption{Execution time of MTASA and state-of-the-art approaches over months}\label{fig:mtasa-time}
\end{figure}

In terms of time execution, MTASA not only excels in accuracy but also demonstrates remarkable efficiency. Its execution times remain remarkably consistent across various months, averaging at 1340 seconds. This efficiency is in stark contrast to the alternative methods, namely CCAFS and Guimapi et al.'s approach, which exhibit more noticeable variations in execution times, at times reaching as high as 2921 seconds and 3147 seconds, respectively. Throughout the entire experiment, MTASA consistently outperforms in terms of speed, being, on average, 1.95 times faster than CCAFS and 2.18 times faster than the approach reported in \cite{GUIMAPI2022e02056}. This significant speed advantage positions MTASA as an efficient solution for time series similarity assessment, making it particularly well-suited for handling large-scale multivariate time series datasets. The consistent efficiency of MTASA can be attributed to its well-optimized multiprocessing computational processes, specifically designed to manage the complexities of working with extensive time series data. Furthermore, to ensure fairness in our comparison, we included the performance of MTASA's sequential version in Figure \ref{fig:mtasa-time}. Even in its sequential form, MTASA proves to be, on average, 1.34 times faster than CCAFS and 1.5 times faster than Guimapi et al.'s approach.

\subsection{Discussion}

Numerous integrated approaches for time series similarity assessment, in literature: \cite{GULLO2012344, Wang2014, Yin2014GeneralizedFF}, and \cite{Liu2023}, often rely on classification techniques. These techniques typically involve clustering time series data in a training dataset using a similarity measure and subsequently associating the query sequence with one of the learned clusters, as explained in \cite{Yoon2016}. While these classification-based methods have demonstrated effectiveness and efficiency in various applications, they come with their own set of limitations. One primary concern is that many of these methods require prior knowledge or information about the number of clusters needed for the classification process \citep{CHANG20101346, LIU20111267}. This requirement may not always be feasible or realistic particularly when dealing with real-world datasets in fields like ecology , where our understanding of ecological phenomena is still evolving, as noted in \cite{Currie2019}. Similarly, in the medical field, some diseases may have limited domain-specific knowledge available, such as known symptoms, making it challenging to apply traditional classification approaches effectively. Furthermore, obtaining domain-specific knowledge for feature selection and representation can also be a challenging task in many cases. The need for such domain-specific information can make these classification-based methods less adaptable when dealing with limited reference data. As a result, their performance in assessing time series similarity may suffer from reduced accuracy and robustness, as pointed out in \cite{Wang2014}. Herein lies one of the key strengths of MTASA. MTASA does not rely on a predefined number of clusters or extensive domain-specific knowledge for its operation. Instead, it leverages innovative techniques like cross-correlation, convolution, and DFT shifting to capture complex temporal patterns and relationships within time series data. This approach enables MTASA to excel in situations where traditional classification-based methods might struggle due to data limitations or evolving understanding of the underlying phenomena.

To address the challenges outlined previously, certain integrated approaches in time series similarity assessment, as exemplified in \cite{Grinsted2004, RamirezVillegas2011ClimateAF, GUIMAPI2022e02056}, opt for a different strategy. These approaches evaluate the similarity between a query sequence and instances in the dataset directly using a similarity index. This strategy eliminates the need for predefined classes or categories, offering a more flexible and adaptable methodology. By quantifying similarity directly, these techniques accommodate scenarios with limited labeled data or situations where intrinsic grouping within the dataset is uncertain. Consequently, these methods prove particularly suitable when the primary objective is to determine the degree of similarity rather than categorize it. MTASA follows a similar path but enhances it by integrating a multiprocessing engine and a data rotation processor into the process. This integration is designed to streamline and enhance the efficiency of index-based integrated time series similarity assessment approaches. MTASA is engineered to be efficient, accurate, and adaptable, making it an asset in the realm of time series similarity assessment.

The efficiency and accuracy of MTASA have been effectively demonstrated through its application to address a domain-specific issue related to the assessment of FAW population suitability. In this context, MTASA outperforms the index-based integrated time series similarity assessment approaches proposed in \cite{RamirezVillegas2011ClimateAF} and \cite{GUIMAPI2022e02056}. An interesting observation emerges from the examination of the methodology presented in \cite{GUIMAPI2022e02056}, where the lower performances are attributed to the scarcity of reference points used to define the rules. This observation offers a critical insight: In a scenario where well-established reference points are either absent or limited, MTASA emerges as a more robust choice for similarity assessment compared to rule-based modeling. Rule-based modeling heavily relies on the availability of data and domain-specific knowledge to define the rules used in computing the similarity index. The crucial implication here is that MTASA offers adaptability to various contexts without excessive reliance on abundant data and extensive domain-specific expertise. This adaptability makes MTASA an invaluable tool, particularly when dealing with situations where data may be sparse or when the understanding of underlying phenomena is still evolving, reinforcing its significance in the field of time series similarity assessment.

To enhance the usability of MTASA and make it more practical for real-world applications, a promising direction to explore is the development of a Decision Support System (DSS) built upon the foundation of \textit{pymtasa}. Such a DSS would not only improve the accessibility of MTASA but also provide valuable visual representations and recommendations tailored to specific contexts and the similarity index matrix generated by the tool. This DSS could be designed to transform the similarity index matrix into an interactive map for the chosen use case of pest population suitability assessment. This map would serve as a powerful visualization tool, highlighting the regions or locations where the pest population is most likely to be suitable for growth and establishment. Such visual representations can be invaluable for stakeholders, as they provide an intuitive way to grasp the spatial distribution of pest suitability, aiding in decision-making processes. Furthermore, the DSS should offer dynamic recommendations based on both real-time and historical data. By continuously updating information and leveraging historical trends, the system can assist stakeholders in making informed decisions regarding pest control strategies. For instance, it can provide alerts and recommendations when conditions become conducive for pest outbreaks or when intervention measures are necessary. This real-time aspect adds a proactive dimension to pest management, enabling timely responses to potential threats. Additionally, the DSS could be designed to integrate with other relevant data sources, such as crop yield data. This integration would provide a more comprehensive and holistic view of pest dynamics and risk factors. By combining pest suitability information with crop yield data, stakeholders can gain insights into how pest populations may impact agricultural production, allowing for more strategic decision-making in terms of crop selection, planting schedules, and pest management practices. 

\section{Conclusion}
\label{sec:sectionVII}

This paper introduces MTASA and \textit{pymtasa}, which together comprise a methodology and a Python package designed for time series similarity assessment. MTASA seamlessly integrates feature representation, similarity measure, and search into a unified and efficient framework. The method leverages fundamental concepts from digital signal processing, including discrete Fourier transform, cross-correlation, convolution, and the DFT shifting theorem, to ensure an effective representation and alignment of time series data. Additionally, MTASA incorporates a multiprocessing engine that optimizes computational processes, reducing execution time and enabling the handling of large-scale time series datasets. The validation of MTASA is demonstrated through its application in an empirical study focused on agroecological similarity assessment. This study showcases MTASA's superiority over state-of-the-art agroecological similarity assessment techniques, achieving twice the efficiency in terms of both accuracy and execution time. Furthermore, MTASA sets itself apart from existing agroecological similarity assessment techniques by consistently delivering results without the need for rigid thresholds, an abundance of reference sites, or domain-specific expertise. In terms of future research, MTASA holds promise as a versatile framework for time series similarity assessment, particularly in complex systems where accounting for temporal shifts is crucial. Its potential applications span a wide range of domains, including climate modeling, financial market prediction, and bioinformatics. Immediate enhancements could involve extending the multiprocessing engine to parallelize tasks related to similarity measurement and search, further optimizing its efficiency. Additionally, developing a decision support system and incorporating additional feature representation techniques and similarity measures would empower researchers and practitioners to explore and apply MTASA more comprehensively across diverse domains and datasets.

\section*{Acknowledgement}
\label{sec:acknowledgements}

The study received financial support from the USAID/ OFDA through the project titled "Reinforcing and Expanding the Community-Based Fall Armyworm Spodoptera frugiperda (Smith) Monitoring, Forecasting for Early Warning and Timely Management to Protect Food Security and Improve Livelihoods of Vulnerable Communities-CBFAMFEW II" grant Number "720FDA20IO00133". Additional funding was obtained from the German Federal Ministry for Economic Cooperation and Development (BMZ), commissioned and administered through the Deutsche Gesellschaft für Internationale Zusammenarbeit (GIZ) Fund for International Agricultural Research (FIA), grant number 18.7860.2–001.00.

\end{document}